\documentclass{article}
\usepackage{spconf,amsmath,graphicx}
\usepackage{epsfig}
\usepackage{amsmath}
\usepackage{amssymb}
\usepackage{amsfonts}    
\usepackage{bm}
\usepackage{url}

\title{A generalized parametric 3D shape representation for\\articulated pose estimation}
\name{Meng Ding and Guoliang Fan\vspace{-0.1in}}
\address{School of Electrical and Computer Engineering\\
Oklahoma State University, Stillwater, OK, USA\\
        \small{meng.ding@okstate.edu; guoliang.fan@okstate.edu}}
\begin{document}
%
\maketitle
\begin{abstract}
We present a novel parametric 3D shape representation, Generalized sum of Gaussians (G-SoG), which is particularly suitable for pose estimation of articulated objects. Compared with the original sum-of-Gaussians (SoG), G-SoG can handle both isotropic and anisotropic Gaussians, leading to a more flexible and adaptable shape representation yet with much fewer anisotropic Gaussians involved. An articulated shape template can be developed by embedding G-SoG in a tree-structured skeleton model to represent an articulated object. We further derive a differentiable similarity function between G-SoG (the template) and SoG (observed data) that can be optimized analytically for efficient pose estimation. The experimental results on a standard human pose estimation dataset show the effectiveness and advantages of G-SoG over the original SoG as well as the promise compared with the recent algorithms that use more complicated shape models.

\end{abstract}
\section{Introduction}
\label{sec:intro}
Shape representation is an important topic in the field of image/video processing and computer vision due to its wide applications. A good shape model not only captures shape variability accurately, but also facilitates the data matching efficiently. One of the most widely used shape models is the mesh surface which can depict the object precisely, but it usually involves a relatively high computational load \cite{Gall_cvpr_2009,mesh_hand_eccv2012}. Some other methods use a collection of geometric primitives, like cylinders or ellipsoids to render the object surface which is compared with the observed shape cues for matching or evaluation \cite{stochastic_tracking,Standford_eccv_2012,geometric_hand_model_cvpr2012}. On the other hand, statistical parametric shape representations become recently popular \cite{metaballs,ellip_point_normal,Stoll_SoG_2d,SoG_hand_iccv2013,jim_wang_1}, particularly in the context of point set registration, where the template is represented by a statistical parametric model \cite{cpd_journal,jian2011GMM} and there is a closed-form expression to optimize the rigid/non-rigid transformation. It is worth noting that the geometric representation and parametric one are closely related but different on the way how the model is involved to compute the cost function during optimization.
\begin{figure}[tb]
\centerline{\psfig{file=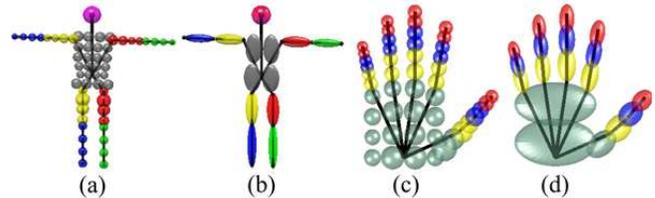, width=3.35in}}
\vspace{-0.1in}
\caption{\small{(a) and (b) show the human shapes with SoG and G-SoG respectively. (c) and (d) are the SoG and G-SoG representations for the hand shape. Obviously, G-SoG has less model order than SoG.}}
\vspace{-0.1in}
\label{fig:all_model}
\end{figure}

Although many 2D/3D shape representations have been proposed for rigid/non-rigid objects, to the best of our knowledge only a few shape representations are amenable to the articulated structure, due to the requirement of an implicit kinematic skeleton. In \cite{Mao_ye_2014,Gall_cvpr_2009}, the detailed mesh model is able to be deformed by the twist-based transformation around the controlling joints for articulated pose estimation. This method could achieve accurate results with a high computational load during the template matching. In \cite{Standford_eccv_2012}, a geometric representation was used to estimate the human pose by an improved Iterated Closest Point (ICP). A sphere-based geometric model was designed for articulated hand representation in \cite{hand_cvpr2014}. As a parametric model, the sum of Gaussians (SoG) model was developed and further embedded in a human skeleton for the articulated pose estimation from multi-view video sequences in \cite{Stoll_SoG_2d}. This simple yet efficient parametric shape representation offers a differentiable model-to-image similarity function, allowing a fast and accurate markerless motion capture system. The SoG representation was further used in \cite{SoG_3d,meng_isvc,SoG_hand_iccv2013} for the human or hand pose estimation.

Inspired by SoG, where only isotropic Gaussians are involved, we develop a more general parametric shape representation, referred to as generalized sum of Gaussians (G-SoG), which can handle both the isotropic and anisotropic Gaussians for fast articulated pose estimation. Compared with the SoG (Fig.~\ref{fig:all_model} (a,c)) in the context of human or hand modeling, G-SoG (Fig.~\ref{fig:all_model} (b,d)) using fewer anisotropic Gaussians, whose volumetric density is modeled by a full covariance matrix, has better flexibility and adaptability to depict various articulated objects, as shown in Fig.~\ref{fig:all_model}. Our G-SoG shares a similar spirit with \cite{SAG_hand_2014} where a sum of anisotropic Gaussians is designed for hand modeling, but we are different in the way that the shape model is used and in terms of the definition of the cost function as well as the implementation of optimization. In \cite{SAG_hand_2014}, each 3D anisotropic Gaussian was projected to a 2D image for an image-to-image similarity evaluation, but our work directly evaluate the model-to-model similarity in the 3D space with an analytical expression that is suitable for the Quasi-Newton optimizer.

 The key contributions of this paper are: (i) We investigate the idea of using G-SoG as a general and simple way to represent articulated objects. (ii) We embed the G-SoG into an articulated skeleton through a tree-structured quaternion-based transformation that enables the gradient-based optimization. (iii) Due to the anisotropic nature, the original similarity function between two SoG models cannot be extended to G-SoG straightforwardly. Thus we derive a new measurement that evaluates the similarity between G-SoG (the shape template) and SoG (the observed data), leading to a closed-form solution to support efficient and accurate pose estimation. The new G-SoG shape representation is simple, flexible and has low computational load, which has a great potential to deal with many articulate pose tracking problems, as shown by the experimental results on a standard depth dataset for 3D human pose estimation.


\section{G-SoG Shape Representations}\label{sec:sec_all2}
We will first introduce the SoG and G-SoG models. Then we present a quaternion-based articulated structure that is embedded with G-SoG for articulated shape representation.

\subsection{SoG and G-SoG}
A single un-normalized 3D Gaussian $G$ has the form,
\begin{equation}\label{equ:intro_sog}
G(\mathbf{x}) = \exp\left(-\frac{||\mathbf{x}-\bm{\mu}||^2}{2\sigma^2}\right),
\end{equation}
where $\mathbf{x}=[x,y,z]^T$ is a vector in the 3D space $\mathbb{R}^3$, $\sigma^2$ and $\bm{\mu} \in \mathbb{R}^3$ are the variance and the mean, respectively. Several spatial Gaussians can be combined as a Sum of Gaussians $\mathcal{K}$ to represent a 3D shape,
\begin{equation} \label{equ:sog_K}
\mathcal{K}(\mathbf{x}) = \sum_{i=1}^n G_i(\mathbf{x}).
\end{equation}
If we extend the variance $\sigma^2$ in (\ref{equ:intro_sog}) to be a full $3 \times 3$ covariance matrix, we will obtain an anisotropic Gaussian as,
\begin{equation}\label{equ:non_iso_sog}
G(\mathbf{x}) = \exp\left(-\frac{1}{2}\!\left(\mathbf{x}-\bm{\mu}\right)^T\!\left[\!
\begin{array}{ccc}
C_{11}\!&\!C_{12}\!&\!C_{13}\\
C_{12}\!&\!C_{22}\!&\!C_{23}\\
C_{13}\!&\!C_{23}\!&\!C_{33}\end{array}
\!\right]\!\left(\mathbf{x}-\bm{\mu}\right)\right),
\end{equation}
which inspires us to construct the G-SoG model for more accurate and flexible 3D articulated shape modeling.
Different with the isotropic Gaussian, the anisotropic Gaussian models its volumetric density by a $3\times3$ covariance matrix, that allows more freedom and flexibility to generally depict some articulated shapes with fewer elongated Gaussian components, particularly for approximating ellipsoidal or cylindrical segments in various articulated objects.

\begin{figure}[tb]
\centerline{\psfig{file=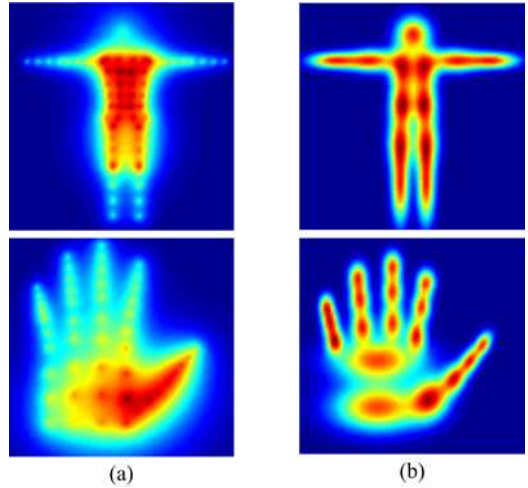, width=2.7in }}
\vspace{-0.1in}
\caption{\small{The volumetric density map comparison of SoG (a) and G-SoG (b) in the projected 2D image. The warmer color means larger density. The variance of each Gaussian in SoG has been manually optimized to obtain a decent color map. Obviously, the silhouette of G-SoG is more distinct, compact and smooth than that of SoG.}}
\vspace{-0.1in}
\label{fig:color_map}
\end{figure}

Fig.~\ref{fig:color_map} exhibits the volumetric density comparison of SoG (a) and G-SoG (b) for a human body and a hand shape in the projected 2D image, where we can observe that the density of G-SoG has a more distinct and compact silhouette than that of SoG, revealing three benefits of G-SoG when approximating an articulated object. First, the density of G-SoG is more smooth and continuous, which facilitates the optimizer to achieve more accurate pose estimation, due to less local minimum in the objective function. Second, the anatomical landmarks (i.e. body/finger joints) have clear definitions in G-SoG. Third, G-SoG has a lower model order (SoG vs. G-SoG, 58:13 in the human body and 48:18 in the hand). Particularly, when the articulated structure is complicated, G-SoG could be much simpler than SoG and has a lower computational cost during the optimization.
\subsection{Quaternion-based Articulated Shape Representation}\label{sec:preliminari}
To have an articulated shape representation, we bind a kinematic skeleton and a G-SoG model $\mathcal{K}_M$ composed by several anisotropic Gaussians, as shown in Fig.~\ref{fig:all_model}(b) and (d). The kinematic skeleton is constructed by a hierarchical tree structure. Each rigid segment in the articulated object (e.g. human body or hand) defines a local coordinate system that can be transformed to the world coordinate system via a $4 \times 4$ matrix $T_l$, as
\begin{equation}\label{equ:transformation}
T_l = T_{par(l)}R_l,
\end{equation}
where $R_l$ is a $4\times4$ matrix containing a pre-defined translation (an offset in the parent's coordinate system) and a 3D rotation formed from the pose parameters. $R_l$ denotes the relative transformation from segment $l$ to its parent $par(l)$.
If $l$ is the root node, $T_{root}$ denotes the global transformation of the whole model. The rotation in each $R_l$ and $T_{root}$ constitute the pose parameters to be estimated. We represent a 3D rotation as a normalized quaternion, which facilitates the gradient-based optimization. If we have $L$ rotations, each of which is represented by a quaternion vector of four elements, we totally have $4\times L$ and $3$ additional global translation parameters in the pose parameters $\mathbf{\Theta}$. Normally, we could attach one anisotropic Gaussian on each elongated segment in the articulated objects, like limbs and fingers, as shown in Fig.~\ref{fig:all_model}(b)(d). Also, we could combine several anisotropic Gaussians to represent one segment, like torso and palm.

\section{Similarity Function of G-SoG and SoG}
To estimate the articulated pose, we derive a new similarity measurement to evaluate the fitness of G-SoG to the observation. Representing observed discrete data points as a continuous density function is necessary for deriving the analytical expression of the similarity function. In this work, we can simply model the observed points as SoG using an uniform variance or via the Octree-based clustering method in \cite{meng_wacv2015,meng_cvprw2015}. Then we use a differentiable similarity function between G-SoG (the template) and SoG (observed data) to optimize pose parameters by an gradient-based optimizer.

\subsection{Similarity between G-SoG and SoG}\label{subsec:gsog_sim}
Given the template $\mathcal{K}_M$ (G-SoG) and the observed data $\mathcal{K}_P$ (SoG), we derive a similarity function to quantify the spatial correlation \cite{correlation_kernel,correlation_eccv} between G-SoG and SoG, which is the integral of the product of $\mathcal{K}_M$ and $\mathcal{K}_P$ over the 3D space $\Omega$,
 \begin{eqnarray}\label{equ:sog_simi_1}
 E(\mathcal{K}_M, \mathcal{K}_P) &=& \int_{\Omega} \sum_{i\in \mathcal{K}_M} \sum_{j\in \mathcal{K}_P}G_i(\mathbf{x})G_j(\mathbf{x})d\mathbf{x}\nonumber\\
 &=& \sum_{i\in \mathcal{K}_M} \sum_{j\in \mathcal{K}_P}E_{ij},
 \end{eqnarray}
 where $E_{ij}$ is the similarity between an anisotropic Gaussian $G_i(\mathbf{x})$ and an isotropic Gaussian $G_j(\mathbf{x})$ that is expanded as,
 \begin{eqnarray}\label{equ:sog_simi_3}
 E_{ij} &=& \iiint_{x,y,z} G_i(x,y,z)G_j(x,y,z)\,dx\,dy\,dz\nonumber\\
&=&
\iiint_{x,y,z}\exp\left\{-\frac{1}{2}\left(\mathbf{A}\right)\right\}\,dx\,dy\,dz,
 \end{eqnarray}
 where,
 \begin{eqnarray}\label{equ:sog_simi_4}
 \mathbf{A}&=&\left[x\!-\!a, y\!-\!b, z\!-\!c\right]\left[\!
\begin{array}{ccc}
C_{11}\!&\!C_{12}\!&\!C_{13},\\
C_{12}\!&\!C_{22}\!&\!C_{23},\\
C_{13}\!&\!C_{23}\!&\!C_{33}\end{array}.
\!\right]\!\left[\!
\begin{array}{c}
\!x\!-\!a\!\\
\!y\!-\!b\!\\
\!z\!-\!c\!\end{array}\!\right]\nonumber\\
&& +\;m(x\!-\!d)^2+m(y\!-\!e)^2+m(z\!-\!f)^2,
 \end{eqnarray}
where vector $[a,b,c]^T$ and $[d,e,f]^T$ are the mean of $G_i$ and $G_j$, respectively. The $3 \times 3$ symmetric matrix \textbf{C} is the covariance matrix of $G_i$, and the scalar $m$ is the reciprocal of the variance in $G_j$. Finally, we have the expression
\begin{equation}\label{equ:sog_simi_5}
E_{ij}= \left(2\pi\right)^{\frac{3}{2}}\frac{1}{|q_{11}q_{22}q_{33}|}\exp\left\{-\frac{1}{2}\left(\mathbf{K}-\mathbf{L}\right)\right\},
\end{equation}
where,
\begin{eqnarray}\label{equ:K}
\mathbf{K}&=&C_{11}a^2 + 2C_{12}ab + 2C_{13}ac + C_{22}b^2 + 2C_{23}bc\nonumber\\
&& +\,C_{33}c^2 + md^2 + me^2 + mf^2,\\
\mathbf{L}&=&q_{14}^2\!+q_{24}^2\!+q_{34}^2.
 \end{eqnarray}
The derivation of (\ref{equ:sog_simi_5}) and the intermediate variables $q_{11}$, $q_{22}$, $q_{33}$, $q_{14}$, $q_{24}$ and $q_{34}$ are detailed in \cite{meng_wacv2015,meng_TIP2016}. Consequently, the similarity $E_{ij}$ is calculated in (\ref{equ:sog_simi_5}) and we obtain the similarity measure $E$ through the sum over $\mathcal{K}_M$ and $\mathcal{K}_P$ via (\ref{equ:sog_simi_1}). Next, we will derive the derivative of our similarity function for the articulated pose estimation.

\subsection{Efficient Articulated Pose Estimation}
The pose parameters are estimated by maximizing the above similarity function. Due to the differentiable nature of the similarity function and the benefits of quaternion-based rotation representation, we can explicitly take the derivative of $E$ with respect to parameter vector $\mathbf{\Theta}$, 
\begin{equation}\label{equ:faiE}
\frac{\partial E(\mathbf{\Theta})}{\partial \mathbf{\Theta}}
=\sum_{i\in\mathcal{K}_M}\sum_{j\in\mathcal{K}_P}\frac{\partial E_{ij}(\mathbf{\Theta})}{\partial \mathbf{\Theta}}.
\end{equation}
We denote $\mathbf{r}=[r_1,r_2,r_3,r_4]^T$ as an un-normalized quaternion, which is normalized to $\mathbf{p}=[x,y,z,w]^T$ according to $\mathbf{p}=\frac{\mathbf{r}}{\|\mathbf{r}\|}$. We expand the pose $\mathbf{\Theta}$ as $[\mathbf{t},\mathbf{r}^{(1)},\ldots,\mathbf{r}^{(L)}]$, where $\mathbf{t}\in \mathbb{R}^3$ defines a global translation and each normalized quaternion $\mathbf{p}^{(l)}$ from $\mathbf{r}^{(l)}\in \mathbb{R}^4$ defines the relative rotation of joint $l$. We denote $\mathbf{u}_i=[a,b,c]^T$ is the mean of $i_{th}$ Gaussian which is transformed from its local coordinate system through transformation $T_i$ in (\ref{equ:transformation}) and the covariance matrix $\mathbf{C}_i$ is approximated from an initial pose or the previous estimated pose. We explicitly expand (\ref{equ:sog_simi_5}) and take derivative with respect to each pose parameter using chain rule:
\begin{eqnarray}
\frac{\partial E_{ij}}{\partial t_n}&=&\frac{\partial E_{ij}}{\partial \mathbf{u}_i}\frac{\partial \mathbf{u}_i}{\partial t_n},~(n=1,2,3) \\
\frac{\partial E_{ij}}{\partial r_m^{(l)}}&=&\frac{\partial E_{ij}}{\partial \mathbf{u}_i}\frac{\partial \mathbf{u}_i}{\partial T_i}\frac{\partial T_i}{\partial \mathbf{p}^{(l)}}\frac{\partial \mathbf{p}^{(l)}}{\partial r_m^{(l)}},~(m=1,...,4)
\end{eqnarray}
which are straightforward to calculate. Alternative to a variant of steepest descent used in \cite{Stoll_SoG_2d,SoG_3d}, we can employ Quasi-Newton optimizer (L-BFGS \cite{lbfgs}) because of its faster convergence. In a practical pose tracking algorithm, the estimated pose in the previous frame could be used as the initialization for the current frame to speed up optimization and to avoid being trapped into a local minima. Additional constraints, such as visibility (occlusion) and pose continuity are helpful to ensure smooth and reliable pose estimation.

\section{Experimental Results}
We evaluate G-SoG and SoG in the context of human pose estimation on a benchmark depth map dataset using the same tracking algorithm \cite{meng_wacv2015,meng_TIP2016}. Also, we compare them with several recent algorithms with different shape models, i.e., the mesh, geometric and parametric ones.

\subsection{Experiment Setup}
Our evaluation is based on the benchmark dataset SMMC-10 \cite{Standford_cvpr_2010}, and we also include a few recent algorithms for comparison. The ground truth data are the 3D marker positions recorded by the optical tracker. We adopt the joint distance error (cm) as the evaluation metric, which measures the average Euclidean distance error between the ground-truth markers and estimated ones over all markers across all frames. It worth noting that because the locations of markers across different body models are different, there exists an inherent and constant displacement that should be subtracted from the error, as a routine in most recent methods. In this paper, we have improved the displacement calculation method used in \cite{meng_wacv2015,meng_TIP2016} by projecting the markers on each segment and computing a local offset for each segment individually.

\subsection{Quantitative Results}
\begin{figure}[!hb]
\centerline{\psfig{file=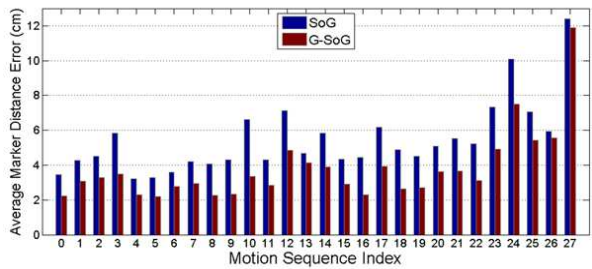, width=3.1in, height = 1.4in}}%
\vspace{-0.15in}
\caption{\small{The accuracy comparison of SoG with G-SoG for human pose estimation in distance error (cm).}}
\label{fig:accuracy_comparison_1}
\vspace{-0.1in}
\end{figure}

\begin{figure}[!htb]
\centerline{\psfig{file=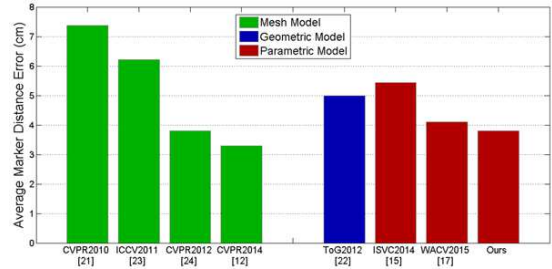, width=2.85in}}
\vspace{-0.1in}
\caption{\small{The accuracy comparison with the state-of-the-art methods \cite{Standford_cvpr_2010,Baak_iccv_2011,taylor_cvpr_2012,Mao_ye_2014,Wei_TAMU_2012,meng_isvc,meng_wacv2015} in distance error (cm).}}
\vspace{-0.1in}
\label{fig:accuracy_comparison_2}
\end{figure}

First, we compare the accuracy of SoG (reported in \cite{meng_isvc}) and G-SoG models under the same real-time tracking algorithm (in C++) \cite{meng_wacv2015} in term of the average joint distance error. In practice, the computational complexity of G-SoG and SoG are comparable in this evaluation. As shown in Fig.~\ref{fig:accuracy_comparison_1}, G-SoG is more accurate than SoG within all 28 sequences. Then, we compare the G-SoG results with several state-of-the-art algorithms in Fig.~\ref{fig:accuracy_comparison_2}, where different shape representations are illustrated in different colors. It worth mentioning that the result of the geometric model in \cite{Wei_TAMU_2012} was tested on a different dataset collected by the authors which is expected to have a better quality (Kinect data $320\times 240$) than the SMMC-10 dataset (SR4000 ToF camera, $176\times 144$). Observing from Fig.~\ref{fig:accuracy_comparison_2}, our G-SoG parametric model can achieve competitive results compared with the best ones using mesh models. Still G-SoG model is simpler, more flexible and efficient than mesh and geometric ones.
\subsection{Qualitative Results}
\begin{figure}[!htb]
\centerline{\psfig{file=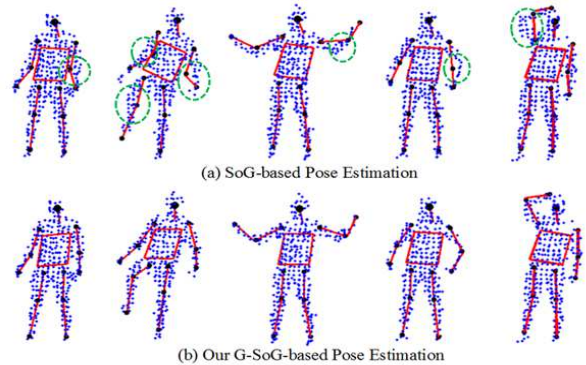, width=3.2in,height = 1.95in}}%
\vspace{-0.1in}
\caption{\small{Comparative analysis of pose estimation: (a) SoG, (b) G-SoG. The green circles in (a) indicate the inaccuracy of SoG results.}}
\label{fig:good_visual_result}
\vspace{-0.1in}
\end{figure}

We visualize some human pose estimation results to compare G-SoG with SoG in Fig.~\ref{fig:good_visual_result}. Obviously, G-SoG achieves more accurate and robust performance (Fig.~\ref{fig:good_visual_result} (b)) compared with SoG (green circles in Fig.~\ref{fig:good_visual_result} (a)). The improvement from G-SoG is likely attributed to the smooth, distinct and compact volumetric density of G-SoG. Especially, the adjustable variance in G-SoG along the depth direction supports better model matching considering the relatively flat depth data.
\vspace{-0.05in}
\section{Conclusion}
We have presented a new G-SoG model for articulated shape representation. G-SoG is more flexible and adaptable than SoG to represent various articulated objects. We have also developed a new differentiable similarity function between G-SoG and SoG that can be optimized analytically to support efficient pose estimation from the depth map or point cloud data. We compare the SoG and G-SoG on a benchmark dataset of human pose estimation. The experimental results show the effectiveness and advantages of G-SoG over SoG as well as its competitiveness with state-of-the-art algorithms. Applications of G-SoG to other articulated objects, e.g., hand pose estimation, industrial VR \cite{Ying_Mao_1,Ying_Mao_2}, medical image registration \cite{Jing_Wu_1,Jing_Wu_2,meng_SPIE_medical}, HCI in mobile device \cite{Lidawei_1,Lidawei_2} and autonomous vehicles \cite{wanhuatong_A_1,wanhuatong_A_2} will be explored in the future.

\vfill
\pagebreak

\bibliographystyle{IEEEbib}

\end{document}